\documentclass[letterpaper]{article} 
\usepackage{aaai24}  
\usepackage{times}  
\usepackage{helvet}  
\usepackage{courier}  
\usepackage[hyphens]{url}  
\usepackage{graphicx} 
\usepackage{natbib}  
\usepackage{caption} 
\usepackage{algorithm}
\usepackage{algorithmic}

\usepackage{amsmath,amsfonts,bm}

\def\eqref#1{equation~\ref{#1}}

\def\1{\bm{1}}

\DeclareMathAlphabet{\mathsfit}{\encodingdefault}{\sfdefault}{m}{sl}
\SetMathAlphabet{\mathsfit}{bold}{\encodingdefault}{\sfdefault}{bx}{n}

\usepackage{multirow}
\usepackage{subcaption}

\usepackage{newfloat}
\usepackage{listings}
\DeclareCaptionStyle{ruled}{labelfont=normalfont,labelsep=colon,strut=off} 
\floatstyle{ruled}
\newfloat{listing}{tb}{lst}{}
\floatname{listing}{Listing}
\title{ATLANTIC: Structure-Aware Retrieval-Augmented Language Model for Interdisciplinary Science}
\author{
    Sai Munikoti, Anurag Acharya, Sridevi Wagle, Sameera Horawalavithana\\
}
\affiliations{
    Pacific Northwest National Laboratory\\

    Richland, WA, USA \\
    \{sai.munikoti,anurag.acharya,sridevi.wagle,\-yasanka.horawalavithana\}@pnnl.gov
}

\usepackage{bibentry}

\begin{document}

\maketitle

\begin{abstract}

Large language models record impressive performance on many natural language processing tasks. 
However, their knowledge capacity is limited to the pretraining corpus. 
Retrieval augmentation offers an effective solution by retrieving context from external knowledge sources to complement the language model. 
However, existing retrieval augmentation techniques ignore the structural relationships between these documents. 
Furthermore, retrieval models are not explored much in scientific tasks, especially in regard to the faithfulness of retrieved documents. 
In this paper, we propose a novel structure-aware retrieval augmented language model that accommodates document structure during retrieval augmentation. 
We create a heterogeneous document graph capturing multiple types of relationships (e.g., citation, co-authorship, etc.) that connect documents from more than 15 scientific disciplines (e.g., Physics, Medicine, Chemistry, etc.).
We train a graph neural network on the curated document graph to act as a structural encoder for the corresponding passages retrieved during the model pretraining. 
Particularly, along with text embeddings of the retrieved passages, we obtain structural embeddings of the documents (passages) and fuse them together before feeding them to the language model. 
We evaluate our model extensively on various scientific benchmarks that include science question-answering and scientific document classification tasks. 
Experimental results demonstrate that structure-aware retrieval improves retrieving more coherent, faithful and contextually relevant passages, while showing a comparable performance in the overall accuracy.

\end{abstract}

\section{Introduction}
The continuous advancement in natural language processing (NLP) has led to the development of various novel model architectures that overcome existing limitations and demonstrate state-of-the-art performances. 
The retrieval augmented language models (RALM) 
primarily address the grounding and scalability challenges in standard language models (LM). 
RALM aims to address these limitations by combining a LM with an external knowledge base. In this framework, the LM generates text conditioned not only on the input query but also on relevant knowledge retrieved from the knowledge base. The retrieved knowledge is usually the text chunks or passages from documents that provide factual grounding to contextualize the model's predictions. In other words, this approach decentralizes model knowledge into parameters and external knowledge sources, thereby addressing the challenges of scalability and adaptability.  

Typically in RALM, text data from an external knowledge base is segmented and encoded into vectors (also known as vector databases). The retriever component of RALM retrieves relevant documents based on the similarity between the query and vectors corresponding to documents in the database. 
Many existing RALMs rely solely on semantic/lexical information of the documents for retrieval. 
However, in certain scenarios, the structural relationship between documents can further support the retriever in retrieving contextually relevant documents.
For instance, a scientific paper in materials science might reference papers that describe relevant advances in nuclear physics, and vice-versa.
Having such relational information explicitly present in the scientific documents would allow the model to draw on the interdisciplinary scientific knowledge in a similar way to how scientists do.
Thus, it would be beneficial to learn about the relationships between documents (e.g., citations, co-authorship, etc.) in a corpus of scientific publications and connect different scientific concepts.


To address the challenges of adequate structural component in RALM and retrieval faithfulness in science-focused tasks, we propose a novel model architecture (ATLANTIC) in this work that systematically incorporates structural and textual information into the RALM. 
We develop ATLANTIC on top of the standard RALM, ATLAS~\citep{izacard2022few} architecture.
In comparison to ATLAS, we introduce new structural encoder component that uses the corresponding structural embeddings along with the text embeddings of the retrieved passages, and then fuse both 
embeddings for each passage before feeding them to the other language modeling components. The structural embeddings are obtained by a pretrained graph neural network on the document relationship graph. This mechanism explicitly incorporates the structural relationship of passages. 
We extensively evaluated the ATLANTIC model on scientific tasks, especially in regard to the faithfulness of the retrieved passages.

\textbf{Contributions} The specific contributions of this work are as follows:
\begin{itemize}
\item Novel mechanism to combine structural and  textual information of scientific documents in the retriever of the RALM model.
\item Structural encoder via a Heterogeneous Graph Transformer (HGT) model pretrained with document relationships.
\item Propose novel evaluation metrics to measure the quality of retrieved documents on scientific tasks. 
\end{itemize}

The rest of the paper is organized as follow. 
Section~\ref{sec:related} provides a brief literature review and Section~\ref{sec:methodology} describes the proposed methodology in details. 
While Section~\ref{sec:ex_setup} outlines the experimental setup, Section~\ref{sec:results} presents the performance analysis and finally Section~\ref{sec:conclusion} concludes the paper.

\section{Related work}
\label{sec:related}
RALM is an active area of research driven by the goal of overcoming the limitations of language models' limited contextual capacity and world knowledge \citep{li2022survey,zhu2023large}.
RALM primarily consists of Retriever (text encoder) and Reader (language model). In the earlier RALM works, retriever is kept frozen and only language model is trained. REALM \citep{guu2020realm} and RAG \citep{lewis2020retrieval} are some of the initial works that focused on retrieving relevant passages from large text corpora to provide additional context to LM. REALM trains an encoder to retrieve passages and pass them to the language model. RAG retrieves documents for Question answering using BM25 and fine-tunes a T5 model along with retrieved passages. Similarly, RETRO \citep{borgeaud2022improving} combines a frozen Bert
retriever, a differentiable encoder and a chunked cross-attention mechanism to predict tokens based on
an order of magnitude more data than what is typically consumed during training. 
REPLUG \citep{shi2023replug}, PKG \citep{luo2023augmented}, and LLM-AMT \citep{wang2023augmenting} propose an alternate plug and play framework where a trainable or even frozen retriever is fused with off-the-shelf frozen language model. DSP \citep{khattab2022demonstrate} provides an in-context learning retrieval augmented framework where retrieved passages act as prompts to the frozen LM. 
HINDSIGHT \citep{paranjape2021hindsight} and ATLAS \citep{izacard2022few} are among few works in the third category where both the retriever and language model are trained in an end to end manner~\citep{hu2023reveal,munikoti2023evaluating,de2023pre}. ATLAS experiments with various designs (in terms of loss functions, pretraining objectives) and training configurations (e.g., query side finetuning vs. full index update) for RALMs with a specific focus on the few-shot learning ability. However, these works solely rely on semantic/lexical information for retrieval augmentation.

In parallel, there are some efforts that looked at incorporating structured knowledge in the form of knowledge graphs. Graph-Retriever~\citep{min2019knowledge} is one of the initial works that iteratively retrieves passages based on the passage relationships, and uses a passage graph to improve passage selection in an extractive reader. KAQA~\citep{zhou2020knowledge} emphasizes improving both document retrieval and candidate answer reranking by considering the relationship between a question and the documents (termed as a question-document graph), and the relationship between candidate documents (termed as a document-document graph). KG-FiD~\citep{yu2021kg} applies KG to a more advanced Fusion in Decoder (FiD) architecture. It uses a graph neural network (GNN) to
re-rank the passages obtained from the retriever and selectively pass a top few for further processing into the LM. 
However, these graph-based RALMs have major shortcomings in terms of (i) accommodating an extra trainable GNN component thereby increasing the computational complexity of the framework; (ii) ranking is not an explicit way of incorporating structural relationships. 

\section{Methodology}
\label{sec:methodology}
\begin{figure*}[!t]
\begin{center}
\includegraphics[width=\textwidth]{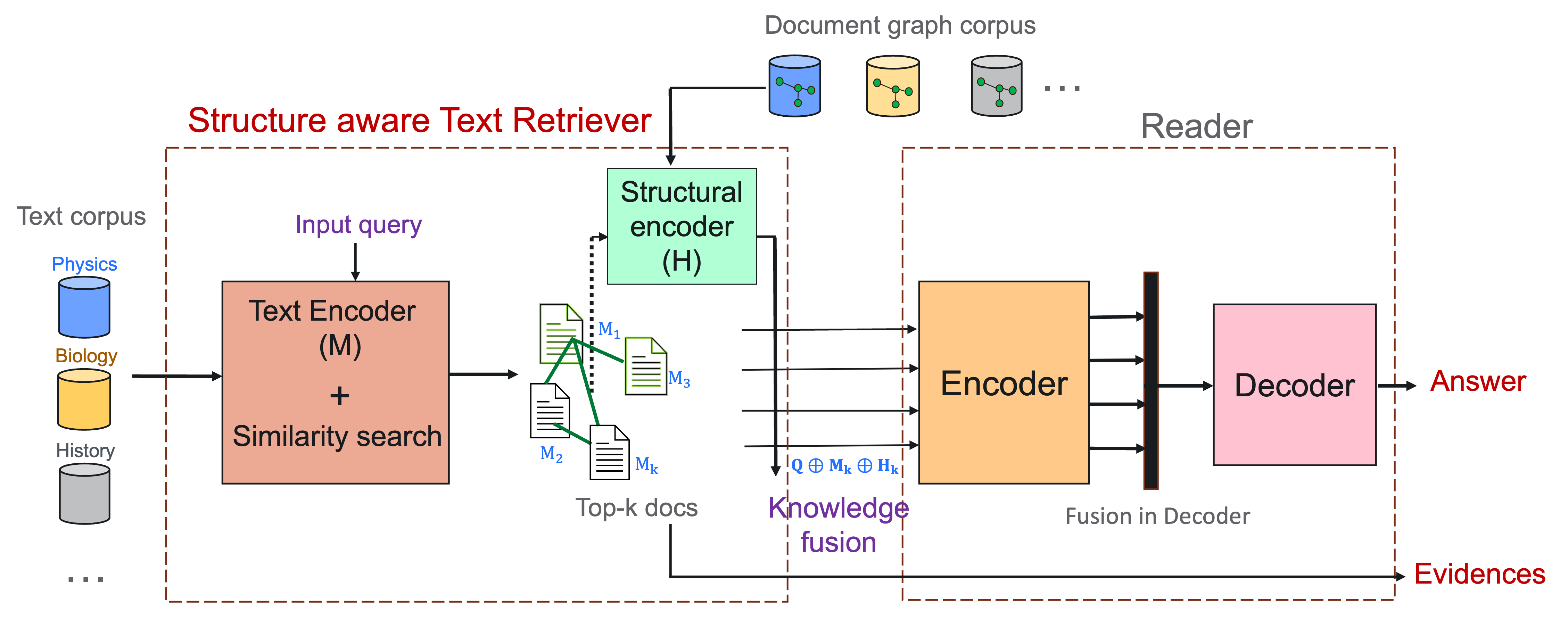}
\end{center}
\caption{Proposed ATLANTIC framework (docs referred to passages). Structural embeddings ($H_{k}$) quantify the cross-document connections among the retrieved docs, which could be useful for multi-hop (multi document) reasoning. For illustration, doc $1$ and doc $3$ are highlighted since their information could offer relatively more relevant context to fetch the answer for the given query.}
\label{fig:Atlantic_architecture}
\end{figure*}

Our approach is based on the ATLAS architecture~\citep{izacard2022few}, which is state-of-the-art RALM. ATLAS consists of a BERT-based \textit{Retriever} model that retrieves top-k passages and feeds along with the input query to the \textit{Reader}, i.e., T5-based LM. 
The basic architecture of our ATLANTIC model is kept the same as that of ATLAS, but we introduced new components and modified the coupling between \textit{Retriever} and \textit{Reader}. 
In ATLANTIC, given the input query, \textit{Retriever} retrieves top-k passages from the input text corpus based on semantic relationship. Unlike ATLAS, which directly passes these top-k passages to the LM, we obtain their structural encodings (embeddings) by leveraging their structural relationships. The structural embeddings are then appended with their semantic counterparts as obtained via \textit{Retriever} encoder, before feeding them to the LM. 
Figure~\ref{fig:Atlantic_architecture} depicts the overview of ATLANTIC architecture with different components and their interactions.
Structural encoding provides extra context to the LM for generation, and it also improves the \textit{Retriever} model to retrieve better passages whose semantic and structural identity better aligns with the target generation. 
Different components of ATLANTIC architecture are described in detail below.

\subsection{Creating Heterogeneous Document Graph}
\label{sec:hdg_create}
Structural encodings for the passages are obtained by leveraging their structural relationships in the form of a Heterogeneous Document Graph (HDG). 
HDG offers plenty of new information that is otherwise ignored in standard semantic-only RALM. 
We first construct the document graph for the text corpus using existing relational information. 
Documents act as nodes, and the relationship between the documents act as links. 
Since our focus is in the scientific domain, we choose a text corpus of scientific articles (research papers), where four kinds of links exist. The link types are co-citation, co-topic, co-venue, and co-institutions where the co-citation links are the majority. 
If document $A$ cites document $B$, then they are connected via a co-citation link. Similarly, if documents belong to the same topic, there is a co-topic link connection. Co-venue and co-institute are applicable when two documents belong to the same publication venue and an institute, respectively. There is a one-to-one mapping between the document and the node in the document graph.

Following earlier works~\citep{karpukhin2020dense,yu2021kg}, each document (article) in our text corpus is split into various disjoint text chunks of $100$ words or $512$ token. Each chunk is called a \textit{passage}, which are fundamental retrieval units. The relationship among the passages is formed based on document-level relations, i.e., if document A and document B are connected via co-topic, then we assume all passages from document A are connected with that of document B. This is achieved by associating all the passages of a document with the same embedding, i.e., their document embedding, so that passages from related documents share similar representation.

\subsection{Text and structure fused knowledge augmentation}
This section describes the proposed framework of fusing text and structural information in the retriever which extracts contextually relevant passages from the external document corpus.

\subsubsection{Text-based Retrieval}
Similar to ATLAS, the retriever in ATLANTIC is based on the Contriever \citep{izacard2021unsupervised}, which retrieve documents based on continuous dense embeddings. It uses dual encoder architecture so that query and passages are embedded independently by a transformer encoder \citep{karpukhin2020dense}. Suppose there are $N$ passages in the text corpus $\{p_{1},p_{2},...,p_{N}\}$, then their embeddings can be represented as:
\begin{equation}
    \pmb{M_{i}}=Contriever(p_{i}) \hspace{0.1cm} \forall \hspace{0.1cm} i \in \{1,2,...,N\},
\end{equation}
where $M_{i}$ belongs to $\mathbb{R}^{D}$ and $D$ is the hidden dimension of the embedding vector. For each input query, the retriever conducts a dot product similarity search between the embedding of the query ($Q$) and embedding of all passages ($M$) as obtained via Contriever, and returns $N_{k}$ passages with the highest similarity scores. Thus, passages are solely retrieved based on their semantic equivalence with the query. $N_{k}$ is substantially smaller than $N$ since we are only interested in extracting a small set of the most relevant passages, typically in the order of tens or hundreds, from the corpus containing millions of passages.

\subsubsection{Structural encoding}
We see in the previous subsection that the retriever model retrieves top $N_{k}$ passages independently based on the semantic/lexical similarity between the query and each passage, and then passes the text embeddings of these $N_{k}$ passages to the LM without accounting for inter-passage relationship. To address this shortcoming, we propose to incorporate the structural relationships via extra embeddings, which we termed as structural embeddings (or encodings). The structural embeddings are then concatenated with text embeddings and passed to the reader (i.e., LM). The structural embeddings are obtained via a structural encoder, which is basically a frozen graph neural network (GNN) model. Particularly, we leverage the heterogeneous graph transformer (HGT) model~\citep{hu2020heterogeneous} to fetch the structural embeddings since they can explicitly account for heterogeneous relationships (co-citations, co-topic, co-venue, co-institute) in the document graph. 
We first train HGT on the document graph using link prediction as the pretraining objective~\citep{hu2020heterogeneous}. The trained HGT is used as a frozen model to encode passages in the ATLANTIC pipeline. The structural embedding of a particular passage $p_{i}$ as obtained from HGT can be represented as:
\begin{equation}
\pmb{H_{i}} = HGT(p_{i}) \hspace{0.1cm} \forall \hspace{0.1cm} i \in \{1,2,...,N_{k}\}.  
\end{equation}
$H_{i}$ is the output from the final encoding layer of the HGT, which basically aggregates the essential information from the neighbors of $p_{i}^{th}$ passage  in the retrieved set $\{p_{1},p_{2},...,p_{N_{k}}\}$. 
This mechanism efficiently utilizes the structural relationships from the document graph. All of the passages belong to a document share the same embedding, i.e., their document embedding as obtained by pretrained HGT. 
It is worth noting that one can use any GNN model to fetch structural embeddings, and the HGT used over here is for illustration purpose.

\subsubsection{Knowledge Fusion}
The semantic retriever and structural encoder provide text embeddings ($M$) and structural embeddings ($H$) of top-k retrieved passages, respectively. We concatenate text and structural embeddings for each passage to generate new aggregate embedding $E_{i}$ as shown below:
\begin{equation}
\begin{split}
\pmb{E_{i}} & = \pmb{M_{i}} \oplus \pmb{H_{i}} \hspace{0.1cm} \forall \hspace{0.1cm} i \in \{1,2,...,N_{k}\},
\end{split}
\end{equation}
where $\oplus$ stands for concatenation operator. The aggregate embedding $\pmb{E}$ captures semantic as well as structural information that enable models to retrieve knowledge from multiple interdisciplinary documents.
$\pmb{E^{'}}$ is the final embedding, which is an input for the reader model. It is obtained by concatenating query embedding $\pmb{Q}$ with the aggregate embedding of the passages $\pmb{E}$ as shown below:
\begin{equation}
\begin{split}
\pmb{E_{i}^{'}} & = \pmb{Q} \oplus \pmb{E_{i}}.
\end{split}
\end{equation}
Since our framework leverages the frozen pretrained GNN model, the novel structural encoder will not induce any computational bottleneck, and its computational complexity is equivalent to that of ATLAS architecture. 

\subsection{Pretraining objectives and loss function}
The retriever and reader (LM) model in ATLANTIC are trained end to end using Perplexity distillation as the loss functions \citep{izacard2020distilling,singh2021end}. The retriever gets feedback from the output of the LM in terms of perplexity score such that it should pick such passages with respect to input query and their structural relationship, which eventually improves the LM perplexity scores.
In this regard, KL-divergence
between the passages distribution of the retriever and the passages posterior distribution is minimized. The loss function can be written as:
\begin{equation}
    L_{i} = \frac{exp(\log p_{LM}(\pmb{a}|\pmb{E_{i}^{'}}))}{\sum_{i=1}^{k}exp(\log p_{LM}(\pmb{a}|\pmb{E_{i}^{'}))}},
\end{equation}
where $\pmb{a}$ denotes the perplexity score of the LM and $p_{LM}$ is the likelihood.

We employ masked language modeling (MLM) \citep{raffel2020exploring} as a pretraining objective. In the given chunk of $M$ tokens, we sample $m$ spans of an average length of three tokens, thereby leading to a mean masking ratio of $15\%$. Then we replace the selected span of tokens with an individual sentinel token. During training, the input to the encoder of the LM is the corrupted (masked) sequence, and the target is then the dropped-out tokens delimited by their sentinel tokens (e.g., $<extra\_id\_0>,<extra\_id\_1>,etc.$). The retriever in our ATLANTIC model retrieves passages using the masked query, but replaces the special mask tokens with a mask token supported by the retriever vocabulary \citep{izacard2022few}.

\section{Experimental Setup}
\label{sec:ex_setup}
In this section, we report the experimental setup used to evaluate the ATLANTIC model on science focused benchmarks.
We outline the datasets, baselines, benchmarks, and training details.

\subsection{Datasets}
We focus on evaluating the ATLANTIC model on its ability to understand scientific language and retrieve contextually relevant passages from multiple scientific knowledge sources. 
We leverage S2ORC~\citep{lo2019s2orc}, which is a large corpus of curated $31.1M$ English-language scientific papers. 
We preprocess the S2ORC~\cite{lo2019s2orc} dataset to create a collection of $354M$ text passages.
Each passage has a maximum of $512$ tokens, or $100$ words, that are concatenated with the corresponding title of the document the passage belongs to. 
Our text corpus captures 19 different scientific domains from the S2ORC collection, which are as follows: 
Art,
Philosophy,
Political-Science,
Sociology,
Psychology,
Geography,
History,
Business,
Economics,
Geology,
Physics,
Chemistry,
Biology,
Mathematics,
Computer Science,
Engineering,
Environmental science,
Material science,
Medicine.
In regard to structural data, we construct a heterogeneous document graph as described in Section~\ref{sec:hdg_create}. 
Table~\ref{tab:graph_stat} (see Appendix) shows the statistics of the S2ORC knowledge graphs which we used to extract the heterogeneous document graphs for each domain.
We use these graphs to train a structural encoder model for each domain.

\subsection{Baseline Models}
To demonstrate the advantages of RALMs on scientific tasks, we choose T5-lm-adapt model~\citep{raffel2020exploring} as a baseline model, which is a standard LM trained on C4 corpus. 
We took the original ATLAS model as a baseline, which is pretrained with common crawl (CC) and Wikipedia on top of the T5 model. 
In addition to this pretrained ATLAS, we also leverage ATLAS-Science model from scratch with the S2ORC scientific text dataset.
For a fair comparison with ATLAS, we initialize the ATLAS-Science model with the \textit{T5-lm-adapt} model and trained jointly with retrieval model, \textit{Contriever}~\citep{izacard2021unsupervised}.
Table~\ref{tab:model_setup} summarizes the baseline model variants with the details of pretraining data.

\begin{table*}[!t]
\centering
\caption{Summary of different pretraining, instruction tuning and benchmark datasets used across baselines and ATLANTIC models.}
\label{tab:model_setup}
\begin{tabular}{|c|c|cc|cc|cc|}
\hline
\multirow{2}{*}{\textbf{Model}}                                                    & \multirow{2}{*}{\textbf{Modality}}                                            & \multicolumn{2}{c|}{\textbf{Pretraining}}                               & \multicolumn{2}{c|}{\textbf{Instruction Tuning}}                      & \multicolumn{2}{c|}{\textbf{Evaluation}}                       \\ \cline{3-8} 
                                                                                   &                                                                               & \multicolumn{1}{c|}{\textbf{Retrieval corpus}} & \textbf{Data}          & \multicolumn{1}{c|}{\textbf{Retrieval corpus}} & \textbf{Data}        & \multicolumn{1}{c|}{\textbf{Retrieval corpus}} & \textbf{Data} \\ \hline
\multirow{3}{*}{\textbf{T5}}                                                       & \multirow{3}{*}{Text}                                                         & \multicolumn{1}{c|}{\multirow{3}{*}{N/A}}      & \multirow{3}{*}{C4}    & \multicolumn{1}{c|}{\multirow{3}{*}{N/A}}         & \multirow{2}{*}{FOS} & \multicolumn{1}{c|}{\multirow{3}{*}{N/A}}      & FOS           \\ \cline{8-8} 
                                                                                   &                                                                               & \multicolumn{1}{c|}{}                          &                        & \multicolumn{1}{c|}{}                          &                      & \multicolumn{1}{c|}{}                          & MAG           \\ \cline{6-6} \cline{8-8} 
                                                                                   &                                                                               & \multicolumn{1}{c|}{}                          &                        & \multicolumn{1}{c|}{}                          & MMLU                 & \multicolumn{1}{c|}{}                          & MMLU          \\ \hline
\multirow{3}{*}{\textbf{ATLAS}}                                                    & \multirow{3}{*}{Text}                                                         & \multicolumn{1}{c|}{\multirow{3}{*}{CC+Wiki}}  & \multirow{3}{*}{Wiki}  & \multicolumn{1}{c|}{\multirow{3}{*}{S2ORC}}    & \multirow{2}{*}{FOS} & \multicolumn{1}{c|}{\multirow{3}{*}{S2ORC}}    & FOS           \\ \cline{8-8} 
                                                                                   &                                                                               & \multicolumn{1}{c|}{}                          &                        & \multicolumn{1}{c|}{}                          &                      & \multicolumn{1}{c|}{}                          & MAG           \\ \cline{6-6} \cline{8-8} 
                                                                                   &                                                                               & \multicolumn{1}{c|}{}                          &                        & \multicolumn{1}{c|}{}                          & MMLU                 & \multicolumn{1}{c|}{}                          & MMLU          \\ \hline
\multirow{3}{*}{\textbf{\begin{tabular}[c]{@{}c@{}}ATLAS-\\ Science\end{tabular}}} & \multirow{3}{*}{Text}                                                         & \multicolumn{1}{c|}{\multirow{3}{*}{S2ORC}}    & \multirow{3}{*}{S2ORC} & \multicolumn{1}{c|}{\multirow{3}{*}{S2ORC}}    & \multirow{2}{*}{FOS} & \multicolumn{1}{c|}{\multirow{3}{*}{S2ORC}}    & FOS           \\ \cline{8-8} 
                                                                                   &                                                                               & \multicolumn{1}{c|}{}                          &                        & \multicolumn{1}{c|}{}                          &                      & \multicolumn{1}{c|}{}                          & MAG           \\ \cline{6-6} \cline{8-8} 
                                                                                   &                                                                               & \multicolumn{1}{c|}{}                          &                        & \multicolumn{1}{c|}{}                          & MMLU                 & \multicolumn{1}{c|}{}                          & MMLU          \\ \hline
\multirow{3}{*}{\textbf{ATLANTIC}}                                                 & \multirow{3}{*}{\begin{tabular}[c]{@{}c@{}}Text\\ +\\ Structure\end{tabular}} & \multicolumn{1}{c|}{\multirow{3}{*}{S2ORC}}    & \multirow{3}{*}{S2ORC} & \multicolumn{1}{c|}{\multirow{3}{*}{S2ORC}}    & \multirow{2}{*}{FOS} & \multicolumn{1}{c|}{\multirow{3}{*}{S2ORC}}    & FOS           \\ \cline{8-8} 
                                                                                   &                                                                               & \multicolumn{1}{c|}{}                          &                        & \multicolumn{1}{c|}{}                          &                      & \multicolumn{1}{c|}{}                          & MAG           \\ \cline{6-6} \cline{8-8} 
                                                                                   &                                                                               & \multicolumn{1}{c|}{}                          &                        & \multicolumn{1}{c|}{}                          & MMLU                 & \multicolumn{1}{c|}{}                          & MMLU          \\ \hline
\end{tabular}
\vspace{-1em}
\end{table*}

\subsection{Benchmarks}
We use two different kinds of scientific benchmarks for training (and finetuning) and evaluating the models. The first benchmark is the SciRepEval~\citep{singh2022scirepeval}  
which provides 25 challenging tasks across four formats: classification, regression, ranking, and search.
In this work, we focus on the classification formatted tasks, \textit{Fields of study (FoS)} and \textit{MAG} due to two main reasons.
First, we need benchmark tasks that test the ability of the models to understand diverse scientific domains and disciplines. FoS tasks include instructions from several disciplines involving existing S2ORC domains as well as new ones. 
For instance, FoS task tests the ability of the model to recognize which domain the given text passage belongs to.
Second, we want to evaluate on specific instruction template to avoid any prompting bias.

Our second evaluation benchmark is MMLU~\citep{hendrycks2020measuring}, which contains $57$ multi-choice question answering datasets (domains) obtained from real examinations designed for humans. 
These datasets cover a wide range of science topics, including high school science, law, and medicine. 
They are broadly categorized into four subsets: humanities, social sciences, STEM, and “other”. We focus on few-shot learning, which leverages 5 training examples per domain. Along with the 5-shot examples, we also leverage additional training examples from other multiple-choice QA tasks provided by the MMLU authors, namely MCTest~\citep{richardson2013mctest}, RACE~\citep{lai2017race}, ARC~\citep{clark2018think} leading to $95k$ training and $14k$ testing examples.

\subsection{Training details}
For training, we create our text corpus and document graph as described earlier. We provide the collection of $354M$ scientific text passages as an external text retrieval corpus for all our finetuning and evaluation tasks. In this regard, we encode all the text passages with the \textit{Contriever} model and construct a document index in the FLAT~\citep{izacard2022few} mode for faster retrieval. 
Retrieval requires frequent updates to the embeddings correspond to the retrieved documents.
However, this update is costly given the size of the retrieval corpus.
To address these scalability issues, we opt for \textit{query side finetuning} approach, which was originally introduced in the ATLAS model~\citep{izacard2022few}. 
This approach is very efficient for model training since it keeps the document encoder frozen while only training the parameters of the query encoder. For a fair comparison, all the models are trained for the same number of tokens. All our experiments are based on base $220M$ model architecture unless explicitly mentioned.

For passage structural embedding, we first train the HGT on the heterogeneous document graph. Thereafter, we obtain the structural embedding of each passage by fetching their respective document encoding via a trained HGT, and consequently saved it in the corresponding index database along with the passage text.
Table~\ref{tab:model_setup} summarizes the pretraining, instruction tuning, and evaluation data used for the baselines and our ATLANTIC model. 
We pretrained the models for $20000$ steps using AdamW as an optimizer with an effective batch size of $32$. We retrieve $20$ passages per query during training. All experiments are conducted on 16 A100 80 GB GPUs in a Linux server. 

We report the performance of the standalone LLM i) T5 (pretrained with C4), ii) ATLAS model (pretrained with CC and Wikipedia), iii) ATLAS-Science  model (pretrained with S2ORC text) and (iv) ATLANTIC (pretrained with S2ORC text and document structure) proposed structural-aware RALM in Section~\ref{sec:results}.

\subsection{Fine Tuning}
\label{sec:Fine tuning}
Previous research~\citep{izacard2022few} has shown that ATLAS model is able to learn knowledge-intensive tasks with very few training examples (i.e., few shot learning).
To allow the model to perform on the scientific downstream tasks, we tune the model with scientific instructions. 
We adopt instruction finetuning for FoS and MAG tasks with a classification style template\footnote{FoS/MAG Instruction Template: \#\#\# Below is an input containing a title-abstract pair. Classify this input into one or more possible Field of Study categories. \#\#\# Possible Categories: [...] \#\#\# Input: \#\# Title: [...] \#\# Response:}.


These templates help guide the model to generate the scientific domain that each passage belongs to. We tune the model with \textit{Fields of study (FoS)} training data after converting them to instructions.
This process resulted in $541,218$ training instructions that were used to perform instruction tuning.
For a fair comparison, we tune all baseline models (T5, ATLAS, ATLAS-Science) with these instructions.
There are $68,147$ and $3,751$ test instructions in the FoS and MAG tasks, respectively. We use MAG instructions to test the out-of-distribution task performance in a zero shot manner.
We followed the same configurations to finetune the model with the MMLU training data as that of ATLAS~\citep{izacard2022few}.

\begin{figure*}[ht]
    \centering
    \begin{subfigure}[t]{0.32\textwidth}
        \centering
        \includegraphics[width=\textwidth]{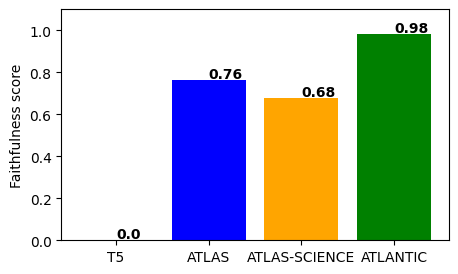}
        \caption{FoS}
        \label{fig:faithfull_FOS}
    \end{subfigure}%
    ~ 
    \begin{subfigure}[t]{0.32\textwidth}
        \centering
        \includegraphics[width=\textwidth]{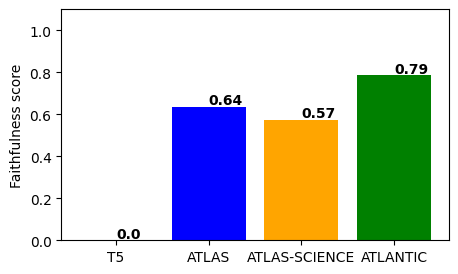}
        \caption{MAG}
        \label{fig:faithfull_MAG}
    \end{subfigure}
    ~ 
    \begin{subfigure}[t]{0.32\textwidth}
        \centering
        \includegraphics[width=\textwidth]{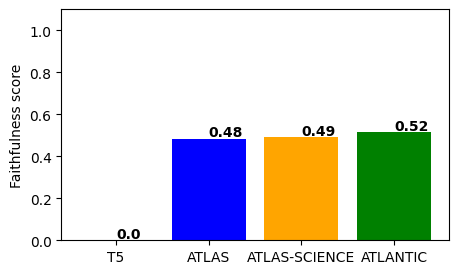}
        \caption{MMLU}
        \label{fig:faithfull_MMLU}
    \end{subfigure}
    \caption{Faithfulness scores across FOS, MAG and MMLU benchmarks. Faithfulness score is the harmonic mean between the accuracy and relevance of the retrieved passages, which gives a holistic view on the trustworthiness of the model. 
    }
    \label{fig:faithfull_fosMag}
\end{figure*}

\subsection{Evaluation metrics}
We use Exact Match (EM) and F1-Score to evaluate the accuracy of generations from RALMs. EM metric evaluates the exact token overlap between the ground truth and generated answers. Furthermore, in existing RALM works, the retriever is mostly evaluated via the generation quality of the language model. 
However, we want to independently evaluate retriever. In this regard, we design two metrics to evaluate the relevance and diversity of the extracted evidences from the retriever:
the \textit{query relevance} and \textit{diversity} metrics. The query relevance metric calculates the semantic similarity of the extracted passages with the input query via their embeddings. Similarity scores are obtained via the dot product of the embeddings. The diversity metric calculates the ratio of the unique evidences in comparison to the total evidences. 
We also devise a new metric, \textit{faithfulness score}, which incorporates the individual performance of both retriever and language model to evaluate the aggregate performance of RALM. Faithfulness score is a measure combining  generation accuracy and relevance score of the retrieved passages via their harmonic mean. It is inspired from F1-Score so that it weights the two metrics (accuracy and relevance score) in a balanced way, requiring both to have a higher value for the faithfulness score value to be high.

\section{Performance Analysis}
\label{sec:results}
In this section, we analyze the performance of ATLANTIC and other baseline models to answer two research questions (RQ 1) and (RQ 2).
We evaluate the performance of models in terms of generation accuracy and quality of the extracted evidences.

\begin{description}
    \item[{(RQ 1)}] Does retrieving structural knowledge help to improve the overall model performance?
    \item[{(RQ 2)}] How useful are the evidences generated from structure-aware RALMs to justify model predictions in science tasks? 
\end{description}

\paragraph{Retrieving structural knowledge helps RALMs to perform better than just retrieving textual
knowledge}
To address (RQ 1), we evaluate the model performance on \textit{Fields of study (FoS)}/\textit{MAG} and \textit{MMLU} benchmarks and compare the performance across ATLANTIC and ATLAS model variants (as shown in Table~\ref{tab:model_setup}). 
For FOS and MAG evaluation, we first finetune all models with only the \textit{FoS} training instructions (as described in Section~\ref{sec:Fine tuning}) and then evaluate on \textit{FoS} (in-distribution) and \textit{MAG} (out-of-distribution) test splits. 
Figures~\ref{fig:faithfull_FOS} and ~\ref{fig:faithfull_MAG} report the performance of the model.
We observe that ATLANTIC model outperforms all other baselines in these benchmarks. 
This indicates that the proposed model has better aggregate performance in terms of retrieving relevant passages and generating correct answers in science tasks. 


\begin{table}[h]
\centering
\caption{Models' ablation study to evaluate performance on MMLU
}
\label{Tab:Results2}
\scalebox{0.95}{
\begin{tabular}{|c|c|c|}
\hline
\textbf{Model} & \textbf{Mean accuracy} & \textbf{Evidence Relevance} \\ \hline
T5             & 0.331                  & N/A                      \\ \hline
ATLAS          & \textbf{0.341}         & 0.825                    \\ \hline
ATLAS-Science  & 0.332                  & 0.928                    \\ \hline
ATLANTIC       & 0.334                  & \textbf{1.135}           \\ \hline
\end{tabular}
}
\end{table}

To further analyze the specific performance of the retriever and reader components, we tabulate their individual results in Table~\ref{Tab:Results1}. 
First, we observe that the accuracy of all RALMs i.e., ATLAS, ATLAS-Science and ATLANTIC are better than that of T5 for both in-distribution \textit{FoS} and out-of-distribution \textit{MAG} tasks, demonstrating the importance of retrieval augmentation. 
Second, we observe that the ATLANTIC ($85.0$ \%) is better than that of ATLAS ($84.40$ \%) and ATLAS-Science ($84.70$ \%) by a small margin of accuracy. 
This demonstrates that retrieving structural knowledge
has low impact in the performance of the reader (language model) for scientific tasks. 
Third, this performance difference is in line with the MMLU benchmark, which we use to evaluate models in science question answering.
To this end, we train all baseline and ATLANTIC models on the MMLU train split similar to the configurations provided in ATLAS~\citep{izacard2022few}. 
We observe a minor difference in the performance of ATLAS and ATLANTIC as shown in Figure~\ref{fig:faithfull_MMLU} and 
Table~\ref{Tab:Results2}.

\begin{table*}[ht]
\centering
\caption{Models' performance on in-distribution (SciDocs-FoS) and out-of-distribution (SciDocs-MAG) benchmarks.}
\label{Tab:Results1}
\begin{tabular}{|c|r|r|r|r|r|r|r|r|}
\hline
\multirow{3}{*}{\textbf{Model}}  & \multicolumn{4}{c|}{\textbf{In-distribution Performance}}                                                                                                                                                                                                                                    & \multicolumn{4}{c|}{\textbf{Out-of-distribution Performance}}                                                                                                                                                                                                                                    \\ \cline{2-9} 
                                &  \multicolumn{2}{c|}{\textbf{Accuracy}} & \multicolumn{2}{c|}{\textbf{Evidence Generation}}                                                                            & \multicolumn{2}{c|}{\textbf{Accuracy}} & \multicolumn{2}{l|}{\textbf{Evidence Generation}}                                                                            \\ \cline{2-9} 
                                & \textbf{EM}     & \textbf{F1}   & \textbf{Relevance} & \textbf{Diversity} & \textbf{EM}     & \textbf{F1}    & \textbf{Relevance} & \textbf{Diversity} \\ \hline
T5                & 0.833                   & 0.87              &                                                                N/A      & N/A                                                                     &   0.579                      &   0.72                  &   N/A                                                                       &    N/A                                                                     \\ \hline
ATLAS       & 0.844                   & \textbf{0.92}             & 0.694                                                                  &      5E-5                                                                  & 0.591                       & \textbf{0.75}                  & 0.69                                                                        & 60E-5                                                 \\ \hline
\textbf{ATLAS-Science}    & 0.847                  & \textbf{0.92}            & 0.564                                                                  &        8E-5                                                                 & 0.578                     & 0.73                    & 0.571                                                                        &      100E-5                                          \\ \hline
\textbf{ATLANTIC}    & \textbf{0.850}                 & 0.89            & \textbf{1.159}                                                                   &        \textbf{10E-5}                                                                 & \textbf{0.595}                     & 0.60                    & \textbf{1.163}                                                                        &      \textbf{120E-5}                                          \\ \hline
\end{tabular}
\end{table*}

\paragraph{Structure aware RALMs retrieve relevant passages to justify model predictions better than text-only models}
To address the (RQ 2), we measure the quality of the retrieved passages across ATLAS, ATLAS-Science and ATLANTIC models.
Though all models perform comparably in the generated answers, they differ significantly in the quality of the passages that they retrieve to support the generated answers.
For example, both ATLAS and ATLAS-Science models achieve low relevance scores in comparison to what achieved by the ATLANTIC model (Table~\ref{Tab:Results1}).
This suggests that the passages retrieved by ATLAS as evidences do not align accurately with the query. 
On the other hand, ATLANTIC retrieves more contextually relevant evidences. 
For example, ATLANTIC model retrieves passages from Chemistry and Biology domains as evidence while ATLAS model retrieves passages from Geology, and Social Science to support a query in Chemistry (see Appendix, Figure~\ref{fig:evidence_traces1}). 

We also analyze the retrieved passes to support MMLU predictions.
As reported in Table \ref{Tab:Results2}, ATLANTIC model records $1.135$ relevance score over $0.825$ in ATLAS.
ATLANTIC retrieves passages from the query domains or at least related domains (such as physics-geology or humanities-social science) whereas ATLAS fails in retrieving passages even from the related domains where the query belongs to (see Appendix, Figure~\ref{fig:MMLU_evidence1} and~\ref{fig:MMLU_evidence2} for sample outputs). 
This suggests that having structured knowledge in the retrieval would help the model to extract most relevant passages to justify model predictions better than the models retrieving only textual knowledge.

\paragraph{Discussions:} With our experiments, we can conclude that ATLANTIC offers better aggregate performance than baseline especially with respect to retriever. One potential explanation for observing minor gain in the language model accuracy despite improved retrieval in ATLANTIC could be that many of the questions in scientific benchmarks (at least the ones we used) are fact-based. For such factual queries, the language model may be less sensitive to the context from retrieved passages, and its more memorizable. The impact of retrieval will be evident in those benchmarks (queries) that are very context dependent. Therefore, even though our model is doing better in terms of aggregate performance due to better retrieval, its impact on the accuracy of the language model is low.
We also urge the scientific community to develop benchmarks that test the ability of the models to perform on interdisciplinary science tasks.

We also noted that some design configurations may have some negative impacts on the effectiveness of the model.
For example, the retrieval corpus was frozen during model training, but the query encoder was allowed to receive the gradient updates 
to address scalability issues~\citep{izacard2022few}. 
This configuration may lead to the model being less able to generalize to scientific data than what was originally tested for general web-quality data. 
It remains as a future work for developing solutions to address this trade-off between scalability and effectiveness of the RALMs.

\section{Conclusion}
\label{sec:conclusion}
In this paper, we present our model, ATLANTIC with a novel framework to integrate document structural knowledge into retrieval-augmented language models. 
To this end, we use a heterogeneous document graph to represent different types of relationships between scientific documents from more than 15 different scientific domains and develop a fusion strategy to combine the text and structure in the knowledge retrieval.
We evaluate our model in multiple scientific benchmarks to test the quality of the retrieved scientific text passages. 
Our experiments demonstrate that retrieving
structural knowledge helps retrieval-augmented language models to perform better overall than 
only retrieving textual knowledge.
Specifically, structural knowledge helps the models to extract more faithful documents as evidence to support the model predictions.
In the future, we will test our model on a wider range of scientific benchmarks and tasks (e.g., hypothesis generation), including those that require knowledge from multiple scientific disciplines.


\section*{Acknowledgements}
This work was supported by the NNSA Office of Defense Nuclear Nonproliferation Research and Development, U.S. Department of Energy, and Pacific Northwest National Laboratory, which is operated by Battelle Memorial Institute for the U.S. Department of Energy under Contract DE-AC05–76RLO1830.
This article has been cleared by PNNL for public release as PNNL-SA-191272.

\bibliography{aaai24}

\begin{thebibliography}{29}
\providecommand{\natexlab}[1]{#1}

\bibitem[{Borgeaud et~al.(2022)Borgeaud, Mensch, Hoffmann, Cai, Rutherford,
  Millican, Van Den~Driessche, Lespiau, Damoc, Clark
  et~al.}]{borgeaud2022improving}
Borgeaud, S.; Mensch, A.; Hoffmann, J.; Cai, T.; Rutherford, E.; Millican, K.;
  Van Den~Driessche, G.~B.; Lespiau, J.-B.; Damoc, B.; Clark, A.; et~al. 2022.
\newblock Improving language models by retrieving from trillions of tokens.
\newblock In \emph{International conference on machine learning}, 2206--2240.
  PMLR.

\bibitem[{Clark et~al.(2018)Clark, Cowhey, Etzioni, Khot, Sabharwal, Schoenick,
  and Tafjord}]{clark2018think}
Clark, P.; Cowhey, I.; Etzioni, O.; Khot, T.; Sabharwal, A.; Schoenick, C.; and
  Tafjord, O. 2018.
\newblock Think you have solved question answering? try arc, the ai2 reasoning
  challenge.
\newblock \emph{arXiv preprint arXiv:1803.05457}.

\bibitem[{De~Jong et~al.(2023)De~Jong, Zemlyanskiy, FitzGerald, Ainslie,
  Sanghai, Sha, and Cohen}]{de2023pre}
De~Jong, M.; Zemlyanskiy, Y.; FitzGerald, N.; Ainslie, J.; Sanghai, S.; Sha,
  F.; and Cohen, W.~W. 2023.
\newblock Pre-computed memory or on-the-fly encoding? A hybrid approach to
  retrieval augmentation makes the most of your compute.
\newblock In \emph{International Conference on Machine Learning}, 7329--7342.
  PMLR.

\bibitem[{Guu et~al.(2020)Guu, Lee, Tung, Pasupat, and Chang}]{guu2020realm}
Guu, K.; Lee, K.; Tung, Z.; Pasupat, P.; and Chang, M.-w. 2020.
\newblock REALM: Retrieval-Augmented Language Model Pre.
\newblock \emph{Training}.

\bibitem[{Hendrycks et~al.(2020)Hendrycks, Burns, Basart, Zou, Mazeika, Song,
  and Steinhardt}]{hendrycks2020measuring}
Hendrycks, D.; Burns, C.; Basart, S.; Zou, A.; Mazeika, M.; Song, D.; and
  Steinhardt, J. 2020.
\newblock Measuring massive multitask language understanding.
\newblock \emph{arXiv preprint arXiv:2009.03300}.

\bibitem[{Hu et~al.(2020)Hu, Dong, Wang, and Sun}]{hu2020heterogeneous}
Hu, Z.; Dong, Y.; Wang, K.; and Sun, Y. 2020.
\newblock Heterogeneous graph transformer.
\newblock In \emph{Proceedings of the web conference 2020}, 2704--2710.

\bibitem[{Hu et~al.(2023)Hu, Iscen, Sun, Wang, Chang, Sun, Schmid, Ross, and
  Fathi}]{hu2023reveal}
Hu, Z.; Iscen, A.; Sun, C.; Wang, Z.; Chang, K.-W.; Sun, Y.; Schmid, C.; Ross,
  D.~A.; and Fathi, A. 2023.
\newblock Reveal: Retrieval-augmented visual-language pre-training with
  multi-source multimodal knowledge memory.
\newblock In \emph{Proceedings of the IEEE/CVF Conference on Computer Vision
  and Pattern Recognition}, 23369--23379.

\bibitem[{Izacard et~al.(2021)Izacard, Caron, Hosseini, Riedel, Bojanowski,
  Joulin, and Grave}]{izacard2021unsupervised}
Izacard, G.; Caron, M.; Hosseini, L.; Riedel, S.; Bojanowski, P.; Joulin, A.;
  and Grave, E. 2021.
\newblock Unsupervised dense information retrieval with contrastive learning.
\newblock \emph{arXiv preprint arXiv:2112.09118}.

\bibitem[{Izacard and Grave(2020)}]{izacard2020distilling}
Izacard, G.; and Grave, E. 2020.
\newblock Distilling knowledge from reader to retriever for question answering.
\newblock \emph{arXiv preprint arXiv:2012.04584}.

\bibitem[{Izacard et~al.(2022)Izacard, Lewis, Lomeli, Hosseini, Petroni,
  Schick, Dwivedi-Yu, Joulin, Riedel, and Grave}]{izacard2022few}
Izacard, G.; Lewis, P.; Lomeli, M.; Hosseini, L.; Petroni, F.; Schick, T.;
  Dwivedi-Yu, J.; Joulin, A.; Riedel, S.; and Grave, E. 2022.
\newblock Few-shot learning with retrieval augmented language models.
\newblock \emph{arXiv preprint arXiv:2208.03299}.

\bibitem[{Karpukhin et~al.(2020)Karpukhin, O{\u{g}}uz, Min, Lewis, Wu, Edunov,
  Chen, and Yih}]{karpukhin2020dense}
Karpukhin, V.; O{\u{g}}uz, B.; Min, S.; Lewis, P.; Wu, L.; Edunov, S.; Chen,
  D.; and Yih, W.-t. 2020.
\newblock Dense passage retrieval for open-domain question answering.
\newblock \emph{arXiv preprint arXiv:2004.04906}.

\bibitem[{Khattab et~al.(2022)Khattab, Santhanam, Li, Hall, Liang, Potts, and
  Zaharia}]{khattab2022demonstrate}
Khattab, O.; Santhanam, K.; Li, X.~L.; Hall, D.; Liang, P.; Potts, C.; and
  Zaharia, M. 2022.
\newblock Demonstrate-Search-Predict: Composing retrieval and language models
  for knowledge-intensive NLP.
\newblock \emph{arXiv preprint arXiv:2212.14024}.

\bibitem[{Lai et~al.(2017)Lai, Xie, Liu, Yang, and Hovy}]{lai2017race}
Lai, G.; Xie, Q.; Liu, H.; Yang, Y.; and Hovy, E. 2017.
\newblock Race: Large-scale reading comprehension dataset from examinations.
\newblock \emph{arXiv preprint arXiv:1704.04683}.

\bibitem[{Lewis et~al.(2020)Lewis, Perez, Piktus, Petroni, Karpukhin, Goyal,
  K{\"u}ttler, Lewis, Yih, Rockt{\"a}schel et~al.}]{lewis2020retrieval}
Lewis, P.; Perez, E.; Piktus, A.; Petroni, F.; Karpukhin, V.; Goyal, N.;
  K{\"u}ttler, H.; Lewis, M.; Yih, W.-t.; Rockt{\"a}schel, T.; et~al. 2020.
\newblock Retrieval-augmented generation for knowledge-intensive nlp tasks.
\newblock \emph{Advances in Neural Information Processing Systems}, 33:
  9459--9474.

\bibitem[{Li et~al.(2022)Li, Su, Cai, Wang, and Liu}]{li2022survey}
Li, H.; Su, Y.; Cai, D.; Wang, Y.; and Liu, L. 2022.
\newblock A survey on retrieval-augmented text generation.
\newblock \emph{arXiv preprint arXiv:2202.01110}.

\bibitem[{Lo et~al.(2019)Lo, Wang, Neumann, Kinney, and Weld}]{lo2019s2orc}
Lo, K.; Wang, L.~L.; Neumann, M.; Kinney, R.; and Weld, D.~S. 2019.
\newblock S2ORC: The semantic scholar open research corpus.
\newblock \emph{arXiv preprint arXiv:1911.02782}.

\bibitem[{Luo et~al.(2023)Luo, Xu, Zhao, Geng, Tao, Ma, Lin, and
  Jiang}]{luo2023augmented}
Luo, Z.; Xu, C.; Zhao, P.; Geng, X.; Tao, C.; Ma, J.; Lin, Q.; and Jiang, D.
  2023.
\newblock Augmented Large Language Models with Parametric Knowledge Guiding.
\newblock \emph{arXiv preprint arXiv:2305.04757}.

\bibitem[{Min et~al.(2019)Min, Chen, Zettlemoyer, and
  Hajishirzi}]{min2019knowledge}
Min, S.; Chen, D.; Zettlemoyer, L.; and Hajishirzi, H. 2019.
\newblock Knowledge guided text retrieval and reading for open domain question
  answering.
\newblock \emph{arXiv preprint arXiv:1911.03868}.

\bibitem[{Munikoti et~al.(2023)Munikoti, Acharya, Wagle, and
  Horawalavithana}]{munikoti2023evaluating}
Munikoti, S.; Acharya, A.; Wagle, S.; and Horawalavithana, S. 2023.
\newblock Evaluating the Effectiveness of Retrieval-Augmented Large Language
  Models in Scientific Document Reasoning.
\newblock \emph{arXiv preprint arXiv:2311.04348}.

\bibitem[{Paranjape et~al.(2021)Paranjape, Khattab, Potts, Zaharia, and
  Manning}]{paranjape2021hindsight}
Paranjape, A.; Khattab, O.; Potts, C.; Zaharia, M.; and Manning, C.~D. 2021.
\newblock Hindsight: Posterior-guided training of retrievers for improved
  open-ended generation.
\newblock \emph{arXiv preprint arXiv:2110.07752}.

\bibitem[{Raffel et~al.(2020)Raffel, Shazeer, Roberts, Lee, Narang, Matena,
  Zhou, Li, and Liu}]{raffel2020exploring}
Raffel, C.; Shazeer, N.; Roberts, A.; Lee, K.; Narang, S.; Matena, M.; Zhou,
  Y.; Li, W.; and Liu, P.~J. 2020.
\newblock Exploring the limits of transfer learning with a unified text-to-text
  transformer.
\newblock \emph{The Journal of Machine Learning Research}, 21(1): 5485--5551.

\bibitem[{Richardson, Burges, and Renshaw(2013)}]{richardson2013mctest}
Richardson, M.; Burges, C.~J.; and Renshaw, E. 2013.
\newblock Mctest: A challenge dataset for the open-domain machine comprehension
  of text.
\newblock In \emph{Proceedings of the 2013 conference on empirical methods in
  natural language processing}, 193--203.

\bibitem[{Shi et~al.(2023)Shi, Min, Yasunaga, Seo, James, Lewis, Zettlemoyer,
  and Yih}]{shi2023replug}
Shi, W.; Min, S.; Yasunaga, M.; Seo, M.; James, R.; Lewis, M.; Zettlemoyer, L.;
  and Yih, W.-t. 2023.
\newblock Replug: Retrieval-augmented black-box language models.
\newblock \emph{arXiv preprint arXiv:2301.12652}.

\bibitem[{Singh et~al.(2022)Singh, D'Arcy, Cohan, Downey, and
  Feldman}]{singh2022scirepeval}
Singh, A.; D'Arcy, M.; Cohan, A.; Downey, D.; and Feldman, S. 2022.
\newblock SciRepEval: A Multi-Format Benchmark for Scientific Document
  Representations.
\newblock \emph{arXiv preprint arXiv:2211.13308}.

\bibitem[{Singh et~al.(2021)Singh, Reddy, Hamilton, Dyer, and
  Yogatama}]{singh2021end}
Singh, D.; Reddy, S.; Hamilton, W.; Dyer, C.; and Yogatama, D. 2021.
\newblock End-to-end training of multi-document reader and retriever for
  open-domain question answering.
\newblock \emph{Advances in Neural Information Processing Systems}, 34:
  25968--25981.

\bibitem[{Wang, Ma, and Chen(2023)}]{wang2023augmenting}
Wang, Y.; Ma, X.; and Chen, W. 2023.
\newblock Augmenting Black-box LLMs with Medical Textbooks for Clinical
  Question Answering.
\newblock \emph{arXiv preprint arXiv:2309.02233}.

\bibitem[{Yu et~al.(2021)Yu, Zhu, Fang, Yu, Wang, Xu, Ren, Yang, and
  Zeng}]{yu2021kg}
Yu, D.; Zhu, C.; Fang, Y.; Yu, W.; Wang, S.; Xu, Y.; Ren, X.; Yang, Y.; and
  Zeng, M. 2021.
\newblock Kg-fid: Infusing knowledge graph in fusion-in-decoder for open-domain
  question answering.
\newblock \emph{arXiv preprint arXiv:2110.04330}.

\bibitem[{Zhou et~al.(2020)Zhou, Shi, Huang, and Zhu}]{zhou2020knowledge}
Zhou, M.; Shi, Z.; Huang, M.; and Zhu, X. 2020.
\newblock Knowledge-aided open-domain question answering.
\newblock \emph{arXiv preprint arXiv:2006.05244}.

\bibitem[{Zhu et~al.(2023)Zhu, Yuan, Wang, Liu, Liu, Deng, Dou, and
  Wen}]{zhu2023large}
Zhu, Y.; Yuan, H.; Wang, S.; Liu, J.; Liu, W.; Deng, C.; Dou, Z.; and Wen,
  J.-R. 2023.
\newblock Large Language Models for Information Retrieval: A Survey.
\newblock \emph{arXiv preprint arXiv:2308.07107}.

\end{thebibliography}

\newpage
\appendix

\section*{Heterogeneous Document Graph Statistics}
The Table \ref{tab:graph_stat} reports the statistics of the document graph domain wise.

\begin{table}[ht]
\caption{S2ORC Knowledge Graph Statistics.}
\centering
\label{tab:graph_stat}
\begin{tabular}{|l|lll|lll|}
\hline
\multicolumn{1}{|c|}{\multirow{2}{*}{Scientific Domain}} & \multicolumn{3}{c|}{\#Nodes}                                        & \multicolumn{3}{c|}{\#Links}                                                                                \\ \cline{2-7} 
\multicolumn{1}{|c|}{}                                   & \multicolumn{1}{l|}{Papers} & \multicolumn{1}{l|}{Authors} & Venues & \multicolumn{1}{l|}{Paper-writes-Author} & \multicolumn{1}{l|}{Paper-PublishedIn-Venue} & Paper-cites-Paper \\ \hline \hline
Art                                                      & \multicolumn{1}{l|}{1911954}       & \multicolumn{1}{l|}{1084868}        &    \multicolumn{1}{l|}{62244}     & \multicolumn{1}{l|}{2466562}                    & \multicolumn{1}{l|}{293012}                        &      \multicolumn{1}{l|}{67566}                  \\ \hline
Biology                                                  & \multicolumn{1}{l|}{7331543}       & \multicolumn{1}{l|}{6514119}        &     \multicolumn{1}{l|}{53775}      & \multicolumn{1}{l|}{21362778}                    & \multicolumn{1}{l|}{881701}        & \multicolumn{1}{l|}{10080386}                                 \\ \hline
Business& \multicolumn{1}{l|}{3105463}       & \multicolumn{1}{l|}{2392490}        &         \multicolumn{1}{l|}{91310}                    & \multicolumn{1}{l|}{5197775}              & \multicolumn{1}{l|}{536536}   & \multicolumn{1}{l|}{791937}            \\ \hline
Chemistry       & \multicolumn{1}{l|}{9704121}       & \multicolumn{1}{l|}{7605776}        &     \multicolumn{1}{l|}{59378}     & \multicolumn{1}{l|}{31548346}                    & \multicolumn{1}{l|}{833327}                        &        \multicolumn{1}{l|}{2942550}            \\ \hline
Computer Science    & \multicolumn{1}{l|}{10079285}       & \multicolumn{1}{l|}{6700984}   & \multicolumn{1}{l|}{263147}                    & \multicolumn{1}{l|}{25190878}      &  \multicolumn{1}{l|}{4852428}                 &    \multicolumn{1}{l|}{25890139}               \\ \hline
Economics    & \multicolumn{1}{l|}{3259612}       & \multicolumn{1}{l|}{1856345}     & \multicolumn{1}{l|}{58804}                & \multicolumn{1}{l|}{5623986}           & \multicolumn{1}{l|}{291409} &   \multicolumn{1}{l|}{4926950}              \\ \hline
Engineering      & \multicolumn{1}{l|}{8139131}       & \multicolumn{1}{l|}{6489295}        & \multicolumn{1}{l|}{100977}                    & \multicolumn{1}{l|}{18123052} &\multicolumn{1}{l|}{1399082}    &\multicolumn{1}{l|}{3215152}         \\ \hline
Environmental Science  & \multicolumn{1}{l|}{1811696}   & \multicolumn{1}{l|}{2202507}      &\multicolumn{1}{l|}{44863}       & \multicolumn{1}{l|}{4553526}    & \multicolumn{1}{l|}{319323}  &\multicolumn{1}{l|}{136134}                   \\ \hline
Geography       & \multicolumn{1}{l|}{2978349}       & \multicolumn{1}{l|}{2693235}     & \multicolumn{1}{l|}{78629}  
& \multicolumn{1}{l|}{5822277} &     \multicolumn{1}{l|}{394774} & \multicolumn{1}{l|}{221624}             \\ \hline
Geology & \multicolumn{1}{l|}{2729089}       & \multicolumn{1}{l|}{2202712}        & \multicolumn{1}{l|}{30510}   & \multicolumn{1}{l|}{7637245} & \multicolumn{1}{l|}{355179}         & \multicolumn{1}{l|}{4840140}              \\ \hline
History & \multicolumn{1}{l|}{2876722}       & \multicolumn{1}{l|}{1323615}    & \multicolumn{1}{l|}{86920}    
& \multicolumn{1}{l|}{3548622}     & \multicolumn{1}{l|}{374663}  & \multicolumn{1}{l|}{142664}                  \\ \hline
Materials Science       & \multicolumn{1}{l|}{7147352}       & \multicolumn{1}{l|}{6258034}       & \multicolumn{1}{l|}{63829}           & \multicolumn{1}{l|}{22680518}    &   \multicolumn{1}{l|}{1173532} &  \multicolumn{1}{l|}{2718010}    \\ \hline
Mathematics & \multicolumn{1}{l|}{4163967}       & \multicolumn{1}{l|}{2611249}        & \multicolumn{1}{l|}{136469}                    & \multicolumn{1}{l|}{8186619} &    \multicolumn{1}{l|}{635732}&  \multicolumn{1}{l|}{6391187}    \\ \hline
Medicine     & \multicolumn{1}{l|}{28504536}       & \multicolumn{1}{l|}{18078042}        &      \multicolumn{1}{l|}{112321}                    & \multicolumn{1}{l|}{98334667} &  \multicolumn{1}{l|}{17419463} & \multicolumn{1}{l|}{67694337}             \\ \hline
Philosophy      & \multicolumn{1}{l|}{1219530} & \multicolumn{1}{l|}{606965}        & \multicolumn{1}{l|}{55551} 
& \multicolumn{1}{l|}{1418267}                        &  \multicolumn{1}{l|}{204771}  &\multicolumn{1}{l|}{125500}               \\ \hline
Physics  & \multicolumn{1}{l|}{6501506}       & \multicolumn{1}{l|}{3814657}        &    \multicolumn{1}{l|}{420139}     & \multicolumn{1}{l|}{28145181}    & \multicolumn{1}{l|}{1211603}      &      \multicolumn{1}{l|}{4771985}             \\ \hline
Political\_Science      & \multicolumn{1}{l|}{3933917}       & \multicolumn{1}{l|}{2268650}        &   \multicolumn{1}{l|}{126063}     & \multicolumn{1}{l|}{5495180}                    & \multicolumn{1}{l|}{693493}                        &    \multicolumn{1}{l|}{489718}                \\ \hline
Psychology          & \multicolumn{1}{l|}{5144736}       & \multicolumn{1}{l|}{3563474}        &  \multicolumn{1}{l|}{104041}      & \multicolumn{1}{l|}{9731235}                    & \multicolumn{1}{l|}{720134}                        &     \multicolumn{1}{l|}{3618670}               \\ \hline
Sociology                                                & \multicolumn{1}{l|}{3993869}       & \multicolumn{1}{l|}{1982767}        &   \multicolumn{1}{l|}{92229}                    & \multicolumn{1}{l|}{5338489}                        &        \multicolumn{1}{l|}{347194}      & \multicolumn{1}{l|}{1475856}   \\ \hline
\end{tabular}
\end{table}

\onecolumn

\section*{Benchmark Examples}
\label{sec:MMLU_evidences}

\begin{figure*}[ht]
\begin{center}
\includegraphics[width=1\textwidth]{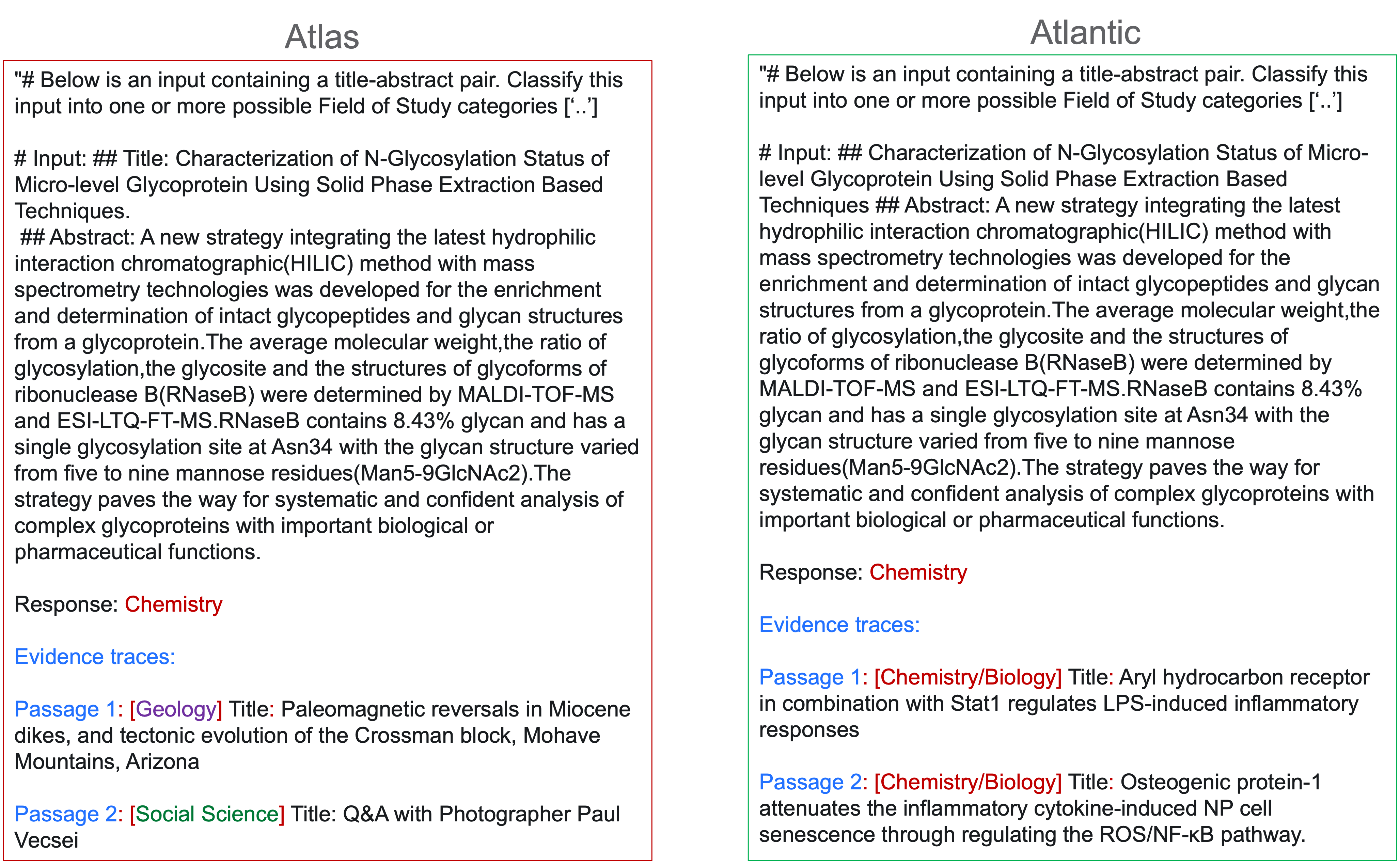}
\end{center}
\caption{Example generations from ATLAS and ATLANTIC models in SciRepEval-FoS~\citep{singh2022scirepeval} task. We color the input query in gray, and the generated answer in red. We list three passages as retrieved by the model to support the answer. We annotate each document by the corresponding scientific domain.}
\label{fig:evidence_traces1}
\vspace{-1em}
\end{figure*}

\begin{figure*}[ht]
\centering
\centering
\includegraphics[width=\linewidth]{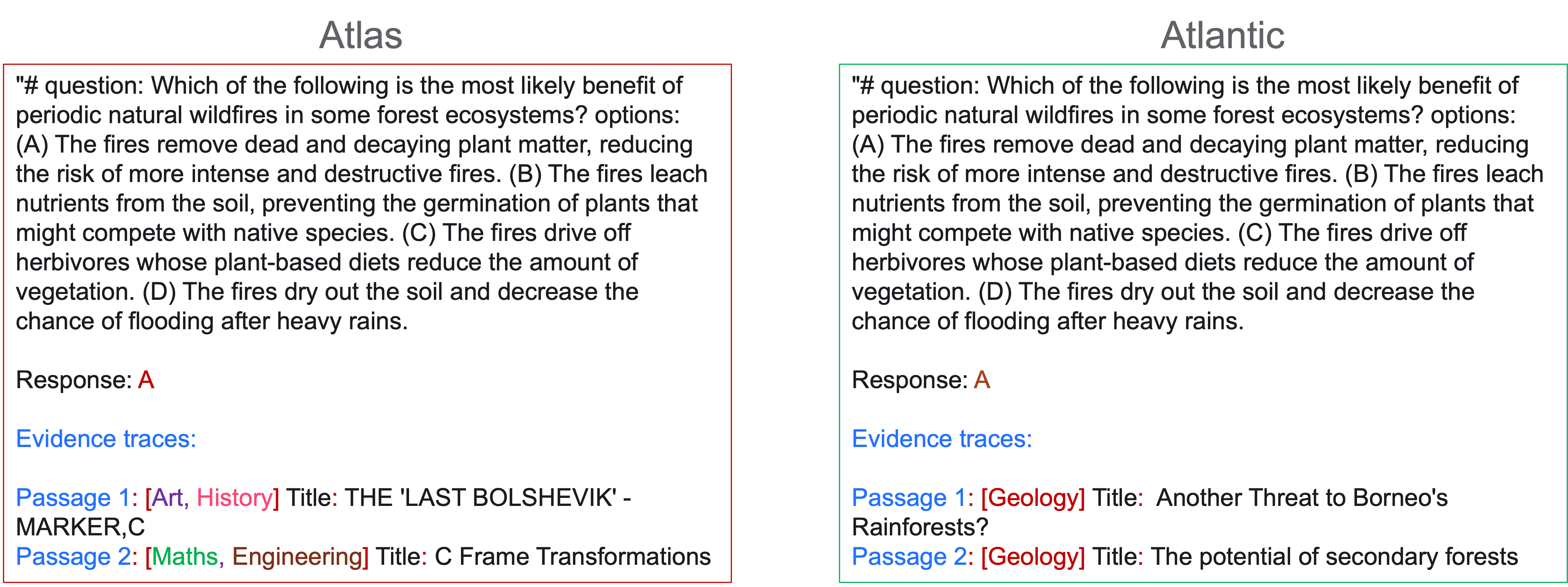}
\centering
\includegraphics[width=\linewidth]{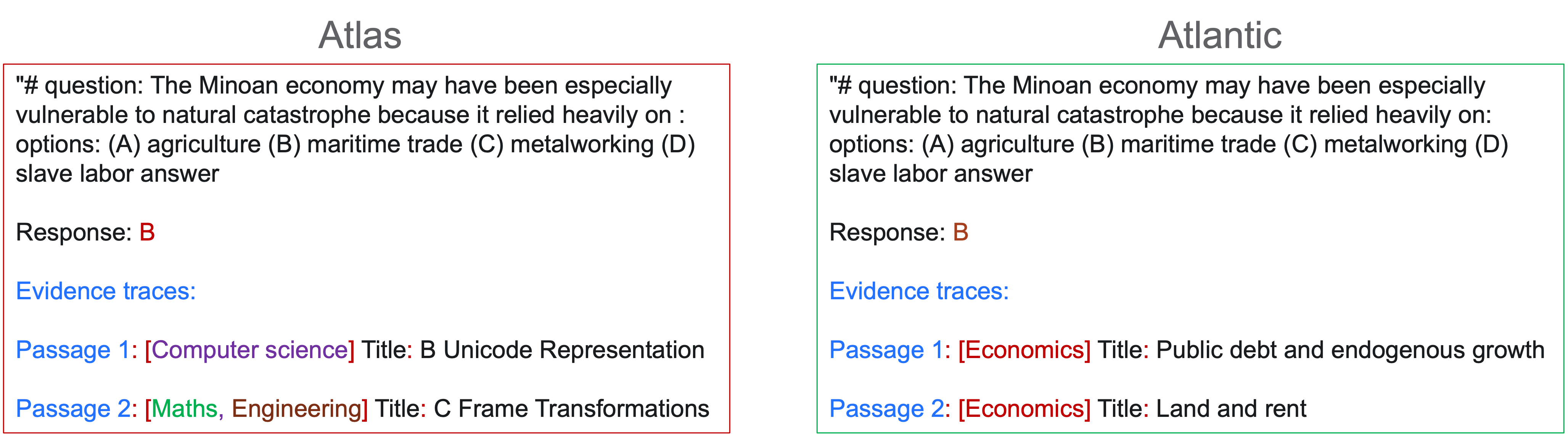}
\caption{Example generations from ATLAS and ATLANTIC models in MMLU task. Accurate retrieval from ATLANTIC but inaccurate retrieval from ATLAS. 
We color the input query in gray, and the generated answer in red. 
We list two passages as retrieved by the model to support the answer. 
We annotate each document by the corresponding scientific domain.}
\label{fig:MMLU_evidence1}
\end{figure*}

\begin{figure*}[ht]
\centering
\centering
\includegraphics[width=\linewidth]{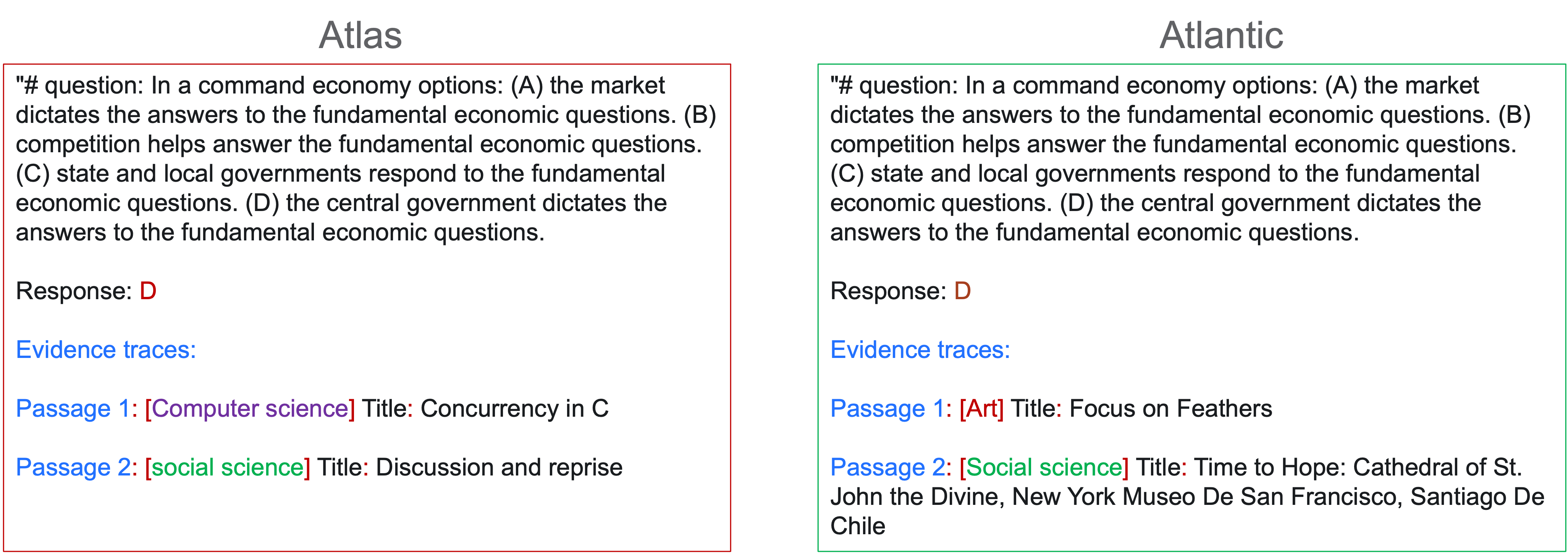}
\centering
\includegraphics[width=\linewidth]{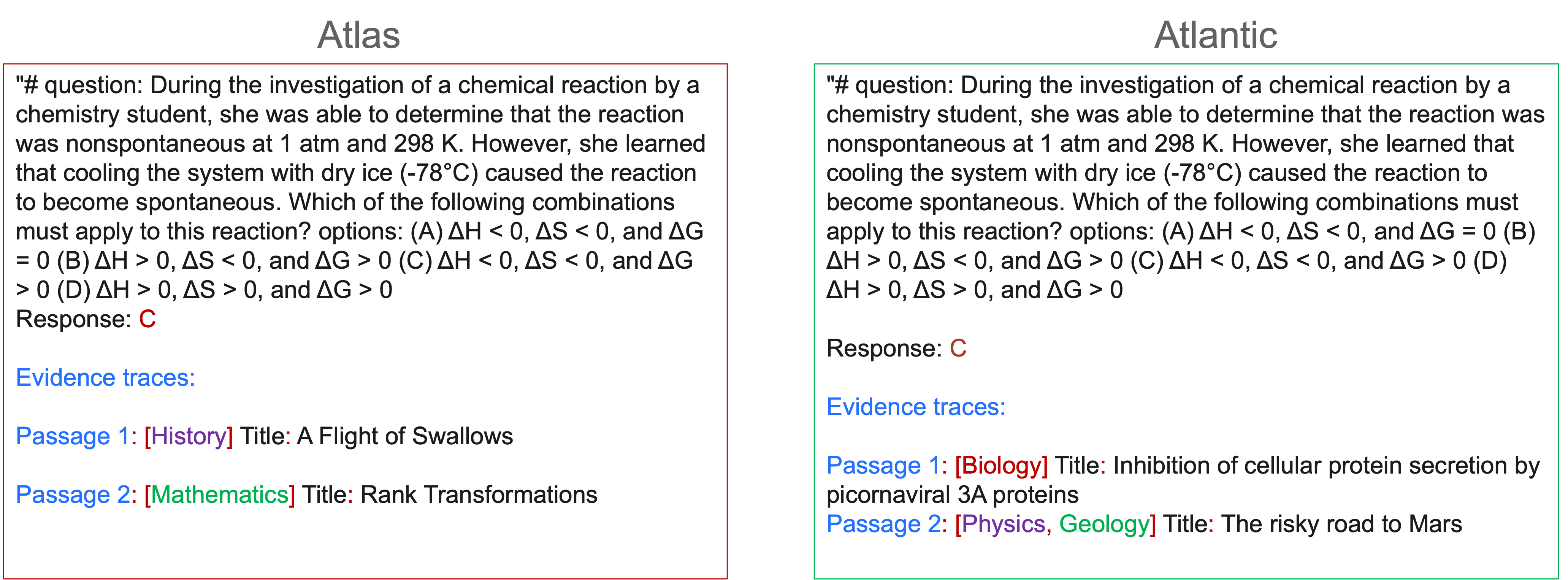}
\caption{Example generations from ATLAS and ATLANTIC models in MMLU benchmark with related inaccurate retrieval to support accurate model predictions. We color the input query in gray, and the generated answer in red. We list two passages as retrieved by the model to support the answer. We annotate each document by the corresponding scientific domain.}
\label{fig:MMLU_evidence2}
\end{figure*}

\end{document}